# Principled Option Learning in Markov Decision Processes

**Roy Fox**[*]  royf@cs.huji.ac.il
**Michal Moshkovitz**[*]  michal.moshkovitz@mail.huji.ac.il
**Naftali Tishby**  tishby@cs.huji.ac.il
*The Hebrew University of Jerusalem*
*Jerusalem, Israel*

## Abstract

It is well known that options can make planning more efficient, among their many benefits. Thus far, algorithms for autonomously discovering a set of useful options were heuristic. Naturally, a principled way of finding a set of useful options may be more promising and insightful. In this paper we suggest a mathematical characterization of good sets of options using tools from information theory. This characterization enables us to find conditions for a set of options to be optimal and an algorithm that outputs a useful set of options and illustrate the proposed algorithm in simulation.

## 1. Introduction

In this paper we consider Markov Decision Processes (MDPs) wherein an agent is confronted with a sequence of subtasks to perform. A subtask may be, for example, to reach a certain region of state-space, starting from the current state. It has been suggested (Sutton et al. (1999); Silver and Ciosek (2012); Mann and Mannor (2013); Ciosek and Silver (2015)) that planning can be made more efficient if a set of useful *options* is available, where each option consists of a policy that solves a simple but relevant subtask. Unfortunately, the problem of discovering useful options autonomously is still open, despite much research effort.

Several heuristic approaches have been proposed for the autonomous discovery of a set of useful options, sometimes also called *option learning*. In one such approach, subtasks were defined using states or regions that the agent tends to visit frequently (McGovern and Barto (2001); Stolle and Precup (2002)). In another approach, properties of the graph defined by the MDP's transition probabilities were used, such as betweenness (Şimşek and Barreto (2009)) or border states of strongly connected regions (Menache et al. (2002)). Others used clustering of the state space (Mannor et al. (2004)). Despite these successes, a principled way of finding a set of useful options may be more promising and insightful.

We present here a mathematical characterization of good sets of options. Intuitively, a good option should perform well in many subtasks, which is guaranteed if it is similar to the optimal solutions of many subtasks. We formalize this notion of similarity using tools from information theory (Tishby and Polani (2011); Rubin et al. (2012)). This characterization enables us to find conditions for a set of options to be optimal.

---

[*]. These authors contributed equally to this work.

Solway et al. (2014) suggest a different mathematical characterization of a useful hierarchy, using the framework of Bayesian model selection. The algorithm they designed requires strong assumptions on the MDP (e.g. the transitions need to be deterministic and reversible). In comparison, our characterization enables us to design an efficient algorithm that applies to any MDP.

In Section 3 below we present our characterization of the learning problem. In Section 4 we then derive the optimality principle and the optimization algorithm. Finally, in Section 5 we illustrate the proposed algorithm in a simulation.

## 2. Preliminaries

### 2.1 Markov Decision Processes

A Markov Decision Process (MDP) is a tuple $(\mathcal{S}, \mathcal{A}, p, c)$ where $\mathcal{S}$ is the state space; $\mathcal{A}$ is the action space; $p\colon \mathcal{S} \times \mathcal{A} \to \Delta(\mathcal{S})$ is a transition distribution function, with $\Delta(\mathcal{S})$ the set of probability distributions over the state space $\mathcal{S}$; and $c\colon \mathcal{S} \times \mathcal{A} \to \mathbb{R}^+$ is a positive cost function. A *policy* is a stochastic mapping from states to actions $\pi\colon \mathcal{S} \to \Delta(\mathcal{A})$. It induces, jointly with the transition distribution $p$, a stochastic process $\mathbb{P}_\pi$ over the trajectory $\rho = (s_0, a_0, s_1, a_1, \ldots, s_T)$ of states and actions, such that in any time step $t$

$$a_t \sim \pi(\cdot|s_t); \qquad s_{t+1} \sim p(\cdot|s_t, a_t).$$

We assume that the episode ends at the first time $T$ that it reaches a terminating state $s_T = s_{\text{term}}$. The agent's objective is to find, for a given initial state $s_0$, a policy that minimizes the total expected cost

$$\mathcal{V}_\pi(s_0) = \mathbb{E}_{\rho \sim \mathbb{P}_\pi(\cdot|s_0)} \left[ \sum_{t=0}^{T-1} c(s_t, a_t) \right]. \tag{1}$$

### 2.2 Options

An option is a policy in a special domain where the agent can choose the halting action $a_{\text{term}}$ that terminates the episode, i.e. $p(s_{\text{term}}|s, a_{\text{term}}) = 1$ for any $s \in \mathcal{S}$. Some researchers also define a set of initial states where the option is allowed to be used (Sutton et al. (1999)), however this part of the definition is unneeded for our purposes.

The target of reducing the total cost requires the agent to trade off short-term costs with long-term costs. This basic property of reinforcement learning has an interesting aspect in the context of options, where the agent trades off the goal value of the state in which it chooses to terminate $g(s) = c(s, a_{\text{term}})$ with the cost accumulated while getting to that state. The state-value function $g(s)$ can be viewed as specifying a *subgoal* for the option to achieve upon termination.

A *subtask* $\theta$ is specified by an initial state $s_0 = s_\theta$ and a subgoal $g_\theta : \mathcal{S} \to \mathbb{R}^+$. Each subtask therefore induces a slightly different MDP, where the terminating components of the cost function $c_\theta(\cdot, a_{\text{term}})$ are the appropriate subgoal.

## 3. Learning Setting

What makes a good option? An option is a pre-learned routine behavior that can be invoked by a high-level controller to solve some subtask. It does not have to be optimal for any particular subtask, but it should be reasonably good for many of the subtasks that are likely to be performed. A good option should therefore be used by many subtasks and it should be similar to the optimal solution for each of these subtasks.

In our setting there are several episodes. At the beginning of each episode, the high-level controller observes the subtask $\theta$ and gives the low-level controller a hint $h$. The option $\pi_h$ now becomes the low-level controller's default behavior, if there is no further control from the high-level controller.

In each time step $t$, the high-level controller can also observe the state $s_t$ and decide on the next action $a_t$, according to some option $\pi_\theta$. However, it incurs cost for causing the low-level controller to diverge from its "uncontrolled" prior behavior $\pi_h$. We measure this intrinsic cost by the Kullback-Leibler (KL) divergence between the distributions over trajectories induced by the actual option used $\pi_\theta$ and by the prior option $\pi_h$

$$\mathcal{I}_{\theta,h}(s_0) = \mathbb{D}[\mathbb{P}_{\pi_\theta} \parallel \mathbb{P}_{\pi_h}] = \mathbb{E}_{\rho \sim \mathbb{P}_{\pi_\theta}(\cdot|s_0)} \left[ \sum_{t=0}^{T-1} \log \frac{\pi_\theta(a_t|s_t)}{\pi_h(a_t|s_t)} \right].$$

Since the actions are taken according to the option $\pi_\theta$, the extrinsic value is now $\mathcal{V}_\theta = \mathcal{V}_{\pi_\theta}$ as in (1). Our optimization problem is to minimize the extrinsic value $\mathcal{V}_\theta$ under the constraint that the expected intrinsic cost $\mathcal{I}_{\theta,h}$ is low, or equivalently, to minimize the KL cost under the constraint that the value is low.

This formulation leads to the problem of minimizing the free energy

$$\mathcal{F}_{\theta,h}(s_0) = \mathcal{I}_{\theta,h}(s_0) + \beta \mathcal{V}_\theta(s_0) = \mathbb{E}_{\rho \sim \mathbb{P}_{\pi_\theta}(\cdot|s_0)} \left[ \sum_{t=0}^{T-1} \left( \log \frac{\pi_\theta(a_t|s_t)}{\pi_h(a_t|s_t)} + \beta c_\theta(s_t, a_t) \right) \right]$$
$$= \sum_{s,a} \mathbb{E}_{\rho \sim \mathbb{P}_{\pi_\theta}(\cdot|s_0)}[\#_{s,a}(\rho)] \left( \log \frac{\pi_\theta(a|s)}{\pi_h(a|s)} + \beta c_\theta(s, a) \right),$$

where $\beta$ is the Lagrange multiplier that trades off the extrinsic and the intrinsic costs and $\#_{s,a}(\rho)$ is the number of occurrences of $s_t = s$, $a_t = a$ in the trajectory $\rho$. Equivalently, this can also be be written recursively as

$$\mathcal{F}_{\theta,h}(s) = \mathbb{E}_{\substack{a \sim \pi_\theta(\cdot|s) \\ s' \sim p(\cdot|s,a)}} \left[ \log \frac{\pi_\theta(a|s)}{\pi_h(a|s)} + \beta c_\theta(s, a) + \mathcal{F}_{\theta,h}(s') \right].$$

The total free energy is

$$\mathcal{F} = \mathbb{E}_{\theta \sim P}[\mathcal{F}_{\theta,q(\theta)}(s_\theta)],$$

where $P$ is the subtask distribution and the hint $h$ is chosen according to some assignment $q : \theta \mapsto h$ of subtasks to hints.

If this assignment is unconstrained, the optimal solution has $\pi_h = \pi_\theta$ for each subtask $\theta$ and $q$ is the identity mapping. This solution fails to generalize to unseen subtasks and is avoided by constraining the assignment in some way. In this paper we consider a hard constraint on the number of prior options.

## 4. Finding Good Options

In this section we derive necessary conditions for the optimal prior options $\{\pi_h\}$, the optimal control options $\{\pi_\theta\}$ and the optimal assignment $q : \theta \mapsto h$ from subtasks to hints. These conditions can serve as iterative update equations, where in each iteration each of these parameters is optimized with the others fixed, as summarized in Algorithm 1.

Clearly, the optimal assignment of a subtask $\theta$ to a hint $h$ is

$$q(\theta) = \arg\min_h \mathcal{F}_{\theta,h}(s_\theta) = \arg\min_h \mathcal{I}_{\theta,h}(s_\theta). \tag{2}$$

To find the optimal control option $\pi_\theta$, for a given subtask $\theta$ and with the prior $\pi_{q(\theta)}$ fixed, we need to minimize the corresponding free energy $\mathcal{F}_{\theta,q(\theta)}(s_\theta)$. This problem of optimal control with KL cost relative to a fixed prior was solved in Rubin et al. (2012), where it was shown that the same free energy function is optimal regardless of the initial state. The optimal free energy and partition function satisfy the self-consistent equations

$$\mathcal{F}_{\theta,h}(s) = -\log \mathcal{Z}_{\theta,h}(s) \tag{3}$$

$$\mathcal{Z}_{\theta,h}(s) = \mathbb{E}_{a \sim \pi_h(\cdot|s)}[\exp(-\beta c_\theta(s,a) - \mathbb{E}_{s' \sim p(\cdot|s,a)}[\mathcal{F}_{\theta,h}(s')])], \tag{4}$$

which can be solved by iterating these updates until convergence. The optimal control option is then

$$\pi_\theta(a|s) \propto \pi_{q(\theta)}(a|s) \exp(-\beta c_\theta(s,a) - \mathbb{E}_{s' \sim p(\cdot|s,a)}[\mathcal{F}_{\theta,q(\theta)}(s')]). \tag{5}$$

To find the optimal prior option $\pi_h$, we define the following Lagrangian

$$\mathcal{L}[\pi_h(\cdot|s)] = \mathcal{F}_{\theta,h}(s) + \lambda_{h,s}\left(\sum_a \pi_h(a|s) - 1\right)$$

and take its derivative for each $h$, $s$ and $a$

$$\frac{\partial \mathcal{L}}{\partial \pi_h(a|s)} = \frac{1}{\pi_h(a|s)} \mathbb{E}_{\theta \sim P}[\delta_{q(\theta)=h} \mathbb{E}_{\rho \sim \mathbb{P}_{\pi_\theta}(\cdot|s_\theta)}[\#_{s,a}(\rho)]] + \lambda_{h,s}.$$

We equate this expression to 0 to get that optimally

$$\pi_h(a|s) = \frac{\mathbb{E}_{\theta \sim P}[\delta_{q(\theta)=h} \mathbb{E}_{\rho \sim \mathbb{P}_{\pi_\theta}(\cdot|s_\theta)}[\#_{s,a}(\rho)]]}{\mathbb{E}_{\theta \sim P}[\delta_{q(\theta)=h} \mathbb{E}_{\rho \sim \mathbb{P}_{\pi_\theta}(\cdot|s_\theta)}[\#_s(\rho)]]} = \mathbb{E}_{\theta \sim Q(\cdot|h,s)}[\pi_\theta(a|s)], \tag{6}$$

where

$$Q(\theta|h,s) \propto P(\theta)\delta_{q(\theta)=h} \mathbb{E}_{\rho \sim \mathbb{P}_{\pi_\theta}(\cdot|s_\theta)}[\#_s(\rho)]$$

is the posterior distribution over the subtasks given the hint and the state. The expected count $C(s_0, s)$ of the occurrences of $s$ in a trajectory starting at $s_0$ satisfies the recursion

$$C(s_0, s) \stackrel{\text{def}}{=} \mathbb{E}_{\rho \sim \mathbb{P}_{\pi_\theta}(\cdot|s_0)}[\#_s(\rho)] = \delta_{s=s_0} + \mathbb{E}_{\substack{a_0 \sim \pi_\theta(\cdot|s_0) \\ s_1 \sim p(\cdot|s_0,a_0)}}[C(s_1, s)].$$

**Algorithm 1:** Find useful options

    **input:** $P, p, c, \beta$
    **initialize:** $\forall \theta, h, s \quad \mathcal{F}_{\theta,h}(s) \leftarrow 0, \quad \pi_h$ randomly
    **repeat**
      **for all** $\theta, h$ **do**
        **repeat**
          $\forall s: \quad \mathcal{Z}_{\theta,h}(s) = \mathbb{E}_{a \sim \pi_h(\cdot|s)}[\exp(-\beta c_\theta(s, a) - \mathbb{E}_{s' \sim p(\cdot|s,a)}[\mathcal{F}_{\theta,h}(s')])]$
          $\forall s: \quad \mathcal{F}_{\theta,h}(s) \leftarrow -\log \mathcal{Z}_{\theta,h}(s)$
        **until** convergence
      **end for**
      $\forall \theta: \quad q(\theta) \leftarrow \arg\min_h \mathcal{F}_{\theta,h}(s_\theta)$
      $\forall \theta, s, a: \quad \pi_\theta(a|s) \leftarrow \frac{1}{\mathcal{Z}_{\theta,q(\theta)}(s)} \pi_{q(\theta)}(a|s) \exp\left(-\beta c_\theta(s,a) - \mathbb{E}_{s' \sim p(\cdot|s,a)}[\mathcal{F}_{\theta,q(\theta)}(s')]\right)$
      $\forall h, s, a: \quad \pi_h(a|s) \leftarrow \mathbb{E}_{\theta \sim Q(\cdot|h,s)}[\pi_\theta(a|s)]$
    **until** convergence
    **return** $\{\pi_h\}, q, \{\pi_\theta\}$

It can be expressed in closed form as

$$C = (I - P_{\pi_\theta})^{-1},$$

with $s_{\text{term}}$ excluded from the transition matrix

$$P_{\pi_\theta}(s, s') = \mathbb{E}_{a \sim \pi_\theta(\cdot|s)}[p(s'|s, a)].$$

In summary, to find the optimal solution, we alternate between a planning step and a clustering step. In the planning step, we apply the update equations (3) and (4) until convergence. In the clustering step, we find the optimal assignment (2) and then get the optimal control option (5). Then the prior options are the average over the appropriate cluster given in (6). We repeat these steps until convergence, as summarized in Algorithm 1.

## 5. Experiments

We experimented with the new algorithm in a two-room domain (Figure 1). This MDP consists of 57 states in a $9 \times 9$ grid, with a narrow corridor connecting the two rooms. There are 9 primitive actions that move the agent deterministically in any horizontal, vertical or diagonal direction or stay. The 10th action, $a_{\text{term}}$, terminates the episode.

The distribution $P(\theta)$ over subtasks chooses uniformly an initial state $s_\theta$ in one room and a goal state $s'_\theta$ in the other room. When $a_{\text{term}}$ is chosen, the agent incurs a square-error goal cost $g_\theta(s) = \|s - s'_\theta\|_2^2$. The agent incurs a fixed cost 1 in each step until it terminates. If the agent attempts to move into a wall, it stays in the same position, and incurs the same cost 1.

This domain has $27^2 = 729$ possible subtasks and it is time consuming to find an optimal policy for each one. Instead, we draw uniformly at random only 30 subtasks and run the algorithm using the empirical uniform distribution over this sample. We would like the algorithm to learn options that are also useful in planning for unseen subtasks.

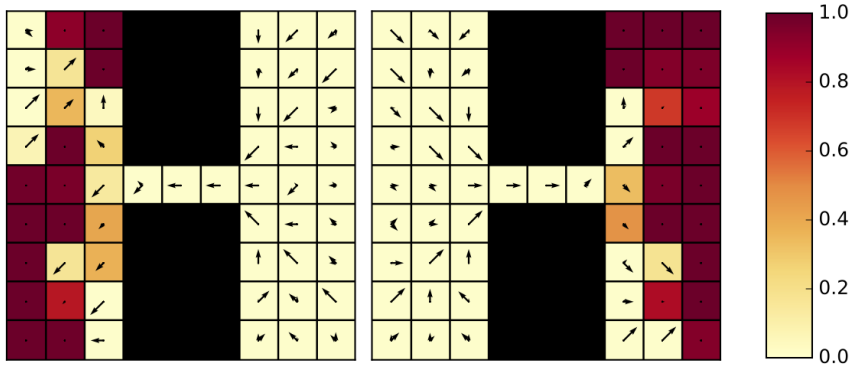

Figure 1: In the two-room domain, the two options returned by our algorithm perform the subtasks of moving to each room.

We used Algorithm 1 to learn 2 prior options for the sampled subtasks, with $\beta = 1$. Each sub-figure of Figure 1 represents a different option $\pi_h$ learned by the algorithm. The lengths of the arrows are proportional to the probability of each move and the color indicates the probability of termination.

Interestingly, each learned prior option is good at performing the subtask of moving to a specific room. The coverage of the state space and the precision of the options would be improved by larger sample sizes, but even with only 30 samples the options generalize well. A policy for an unseen subtask can use one of the options to the correct room, where only local planning is needed to find the correct goal state.

## 6. Conclusions

In this paper we have suggested a mathematical characterization for good sets of options. This characterization follows from the realization that a good option should be similar to the optimal solution of many subtasks. This leads to an optimization criterion that combines two terms — minimizing extrinsic value and minimizing divergence from the option. The latter can be written formally using tools from information theory. This characterization enables us to find conditions for a set of options to be optimal and an algorithm that outputs a useful set of options.

We have used a hard constraint on the number of options, but other constraints can be explored, for example a soft constraint on the conditional entropy of the assignment. Further testing in more complex domains should be conducted in order to fully evaluate the promising ideas presented in this paper.

## Acknowledgments

This work is partially supported by the Gatsby Charitable Foundation, The Israel Science Foundation, and Intel ICRI-CI center. M.M. is grateful to the Harry and Sylvia Hoffman Leadership and Responsibility Program.